# Virtual-point-based Solutions to Handle Generalized Absolute Pose Problem

Bin Li, Banglei Guan, *Member*, *IEEE*, Shunkun Liang, and Yang Shang

***Abstract*—Multi-camera systems are increasingly adopted in robotics and autonomous navigation for their wide field of view, flexibility, and fault tolerance. Nevertheless, existing PnP solvers fail to handle multiple projection centers. This paper introduces a virtual point formulation that bridges the standard PnP and generalized pose problems, enabling a unified pipeline that transforms existing PnP solvers into generalized pose solvers. Based on this framework, we derive three Virtual-point-based Generalized Pose solvers, namely VGPc, VGPq, and VGPr, leveraging Cayley, quaternion, and rotation-matrix parameterizations, respectively. Extensive experiments demonstrate that the proposed solvers inherit the accuracy and efficiency of original PnP algorithms while significantly outperforming existing generalized solvers. Specifically, VGPc achieves higher estimation accuracy under heteroscedastic noise conditions, VGPq maintains global optimality, whereas VGPr provides superior computational efficiency without accuracy degradation. The source code will be made publicly available upon acceptance.**

***Index Terms*—Camera pose estimation, Generalized absolute pose, Multi-camera system, Perspective-n-Point, Virtual point.**

## I. Introduction

VISUAL localization and orientation use images captured by optical imaging systems to determine the camera's spatial position and attitude. By extracting and matching feature points from images, spatial geometric constraints are obtained, enabling the construction of constraint equations to calculate the camera pose at the time of image capture. Consequently, connecting all camera poses allows for the restoration of the complete motion trajectory of the camera and its carrier, assuming the camera is rigidly mounted on the carrier platform. Such capabilities are essential in industrial applications, ranging from robotic manipulation [1] to UAV navigation [2], where accurate localization remains a fundamental challenge. While single-camera systems have been extensively studied [3], multi-camera systems offer distinct advantages, including a wider field of view and fault tolerance [4].

At the core of these systems, visual pose estimation serves as the fundamental computational component of visual localization. Based on feature dimensionality, pose solvers can be categorized into two classes: relative pose solvers (2-D/2-D) and absolute pose solvers (2-D/3-D). The latter requires establishing a reference space (world frame) in advance, extracting the image coordinates of feature points, and matching them to obtain the corresponding 3-D coordinates. Subsequently, constraints from 2-D/3-D point correspondences are used to derive equations for solving for the camera pose. Generally, 2-D/3-D pose estimation is more straightforward and requires fewer points than 2-D/2-D estimation, as one 2-D/3-D point correspondence provides more pose constraints (2 DoF) than one 2-D/2-D correspondence (1 DoF).

*Single-camera absolute pose solvers*: The 2-D/3-D pose estimation problem for a single camera is known as the Perspective-n-Point (PnP) problem, which estimates the camera pose from 2-D/3-D point correspondences. The optimal solution for the 2-D/3-D pose estimation problem is to find the rotation and translation that minimize the reprojection error of feature points. Extensive research has been devoted to developing efficient methods for computing or approximating this optimal solution. These methods can be roughly divided into two categories based on the number of 2-D/3-D point correspondences utilized: minimal solvers and redundant solvers. The former, addressing the P3P problem, requires the fewest correspondences for closed-form solutions [5] and is often combined with the RANSAC algorithm to eliminate mismatches in point correspondences. Conversely, the latter exploits redundant constraints to enhance accuracy and robustness. Under first-order optimality conditions, redundant solvers determine the rotation and translation that minimize specific geometric errors. Specifically, Lepetit et al. [6] introduced four virtual control points to represent 3-D world points and reduced the computational complexity to linear in cases with more than three points. However, the method's accuracy degrades in mildly redundant and planar cases. Hesch et al. [7] proposed a direct least squares (DLS) algorithm to handle all point configurations. The algorithm formulated the PnP problem as a minimization problem, in which the

This work was supported in part by the Hunan Provincial Natural Science Foundation of China under Grant 2026JJ70042, in part by the Science and Technology Innovation Program of Hunan Province under Grant 2025RC1045, and in part by the National Natural Science Foundation of China under Grant 12372189. *(Corresponding author: Yang Shang, Banglei Guan.)*

Bin Li is with the Hunan Institute of Advanced Technology, Changsha 410205, China, and also with the College of Aerospace Science and Engineering, National University of Defense Technology, Changsha 410073, China.

Banglei Guan, Shunkun Liang, and Yang Shang are with the College of Aerospace Science and Engineering, National University of Defense Technology, Changsha 410073, China, and also with the Hunan Provincial Key Laboratory of Image Measurement and Vision Navigation, Changsha 410073, China (e-mail: shangyang1977@nudt.edu.cn, guanbanglei12@nudt.edu.cn).

translation is a linear function of the rotation, parametrized by the Cayley representation. Zheng et al. [8] utilized a non-unit quaternion representation to parameterize rotation and achieved state-of-the-art performance. Nakano et al. [9] derived a new optimality condition for DLS and solved it with Cayley parameters. Lourakis et al. [10] solved the optimality condition derived from Nakano's work with the rotation parameterized by modified Rodrigues parameters. Yu et al. [11] applied fixed pre-rotations to handle singularities of Cayley parameters. These DLS-like or direct minimization methods yield a complicated system of multivariate polynomials that can be solved using Gröbner basis solvers generated by automatic generators [12]. Unlike the aforementioned methods that solve polynomial systems derived from optimality conditions, Terzakis et al. [13] approached the PnP problem as a quadratically constrained quadratic program (QCQP) and obtained a globally optimal solution using a sequential quadratic programming (SQP) scheme. More recently, Vakhitov et al. [14] proposed an uncertainty-aware PnP solver, DLSU. They derived residual covariances that account for both 2-D feature-detection and 3-D reconstruction uncertainties, extending DLS [7] to handle weighted least-squares formulations.

*Generalized absolute pose solvers*: In robotics, augmented reality, aerial localization, and autonomous navigation, camera systems beyond the central projection model are gaining widespread adoption [15]. These cameras, such as stereo, omnidirectional, and multi-camera systems, possess more than one projection center. Among them, multi-camera systems can comprise any number of cameras in arbitrary spatial configurations, providing a broader field of view, greater flexibility, and broader applicability than other camera systems. Because the multi-camera system has multiple projection centers, traditional PnP algorithms designed for a single camera, which assume a single optical center, cannot be directly applied to solve the multi-camera pose problem. To address this issue, several generalized absolute pose solvers have been proposed. Based on the general camera model proposed by Pless et al. [16], Nister et al. [17] introduced a generalized pose estimation method to solve the position and orientation of noncentral cameras, which can be applied to calibrated multi-camera systems. Ventura et al. [18] discussed the generalized pose and scale estimation problem and proposed a corresponding pose solver. Inspired by DLS [7], Kneip et al. [19] proposed UPnP, a generalized pose solver that reformulates the estimation problem as a minimization problem. By solving a polynomial system that depends solely on rotation, UPnP determines the camera position and orientation. This approach was subsequently extended in gDLS [20], [21], which addresses generalized pose and scale estimation under redundant configurations. Notably, gDLS establishes a linear relationship among depth, translation, scale, and rotation variables in its derivation. Combining gDLS and UPnP, Sweeney et al. [22] proposed a faster version, gDLS+++, that uses the same equation-solving strategy as UPnP to accelerate the solution of the generalized pose estimation problem. Subsequently, Wientapper et al. [23] presented an orthogonal-complement formulation that streamlines gDLS's linear parameter elimination via thin-SVD, achieving equivalent accuracy with improved efficiency. gP4Pc [24] utilizes invariants (such as distance ratios) under similarity transformations to establish a polynomial equation system and, adopting the framework of UPnP [19], constructs a minimal solver based on Gröbner basis. gP4Pc computes the depth factors of image rays to determine the coordinates of spatial points in the camera system; subsequently, it solves the 3-D/3-D pose estimation problem to obtain the camera's position and orientation. Thereafter, Liang et al. [25] extended [9] to multi-camera systems by leveraging Lagrange multipliers with Cayley parameterization to solve the generalized absolute pose problem.

However, existing generalized pose solvers typically address the problem in isolation by deriving specific formulations from scratch, rather than leveraging the well-studied PnP solutions. Consequently, the intrinsic relationship between the PnP problem and the generalized pose problem remains underutilized in the current literature. To bridge this gap, this article derives the generalized form of the direct minimization method, transforms the pose estimation problem into a least-squares problem over rotation, and obtains the rotation and translation by solving the resulting optimization problem. Specifically, the main contributions of this article are summarized as follows:

1) *Virtual point representation*: We introduce a virtual point representation that shifts the 3-D point backward along the camera offset. This transformation converts the generalized multi-camera projection equation into the standard single-camera form, implicitly encoding multiple projection centers into the virtual point coordinates and establishing explicit mathematical equivalence between single-camera and multi-camera pose estimation problems.
2) *Unified transformation pipeline*: We present a minimal-modification framework that transforms existing PnP solvers into generalized pose solvers by simply substituting virtual points for the original 3-D points and adjusting the translation-elimination procedure. This pipeline enables the direct inheritance of optimality conditions, convergence properties, and algorithmic implementations from state-of-the-art single-camera methods, without re-deriving the formulations.
3) *Competitive generalized solvers*: By instantiating the framework with Cayley, quaternion, and rotation-matrix parameterizations, we derive VGPc, VGPq, and VGPr. Specifically, VGPc provides uncertainty-aware estimation for high-noise scenarios; VGPq maintains global optimality; and VGPr achieves real-time computation at sub-millisecond execution time. These solvers collectively provide efficient, robust pose estimation in multi-camera applications.

The remainder of this article is organized as follows. Section II reviews the preliminaries and problem formulation. Section III details the proposed virtual point representation and the derivation of the generalized pose solvers. Section IV presents the experimental results on both synthetic and real datasets. Finally, Section V concludes this article.

## II. Fundamental Preliminaries

In this section, the notations and the formulations of single-camera and generalized pose estimation are introduced.

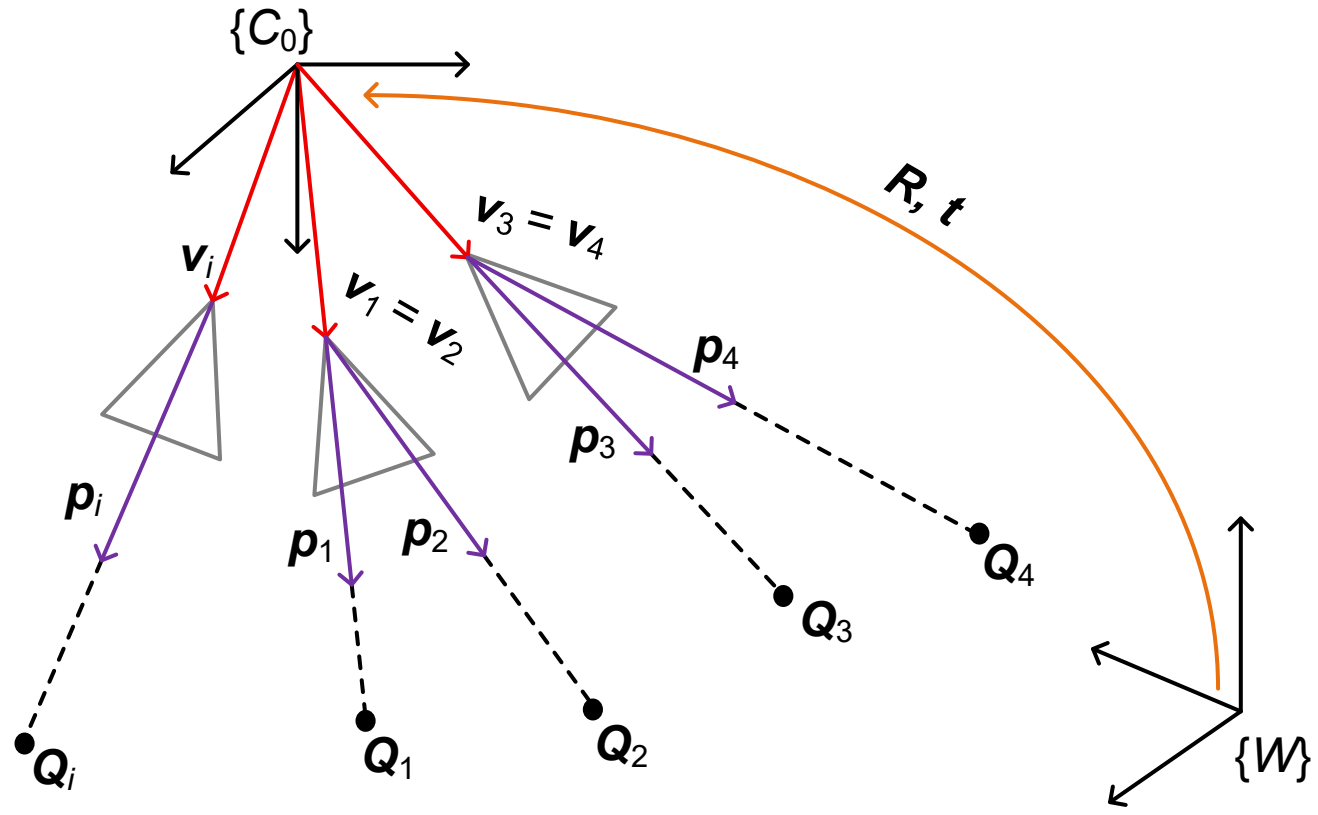

Fig. 1. Projection diagram of generalized camera model.

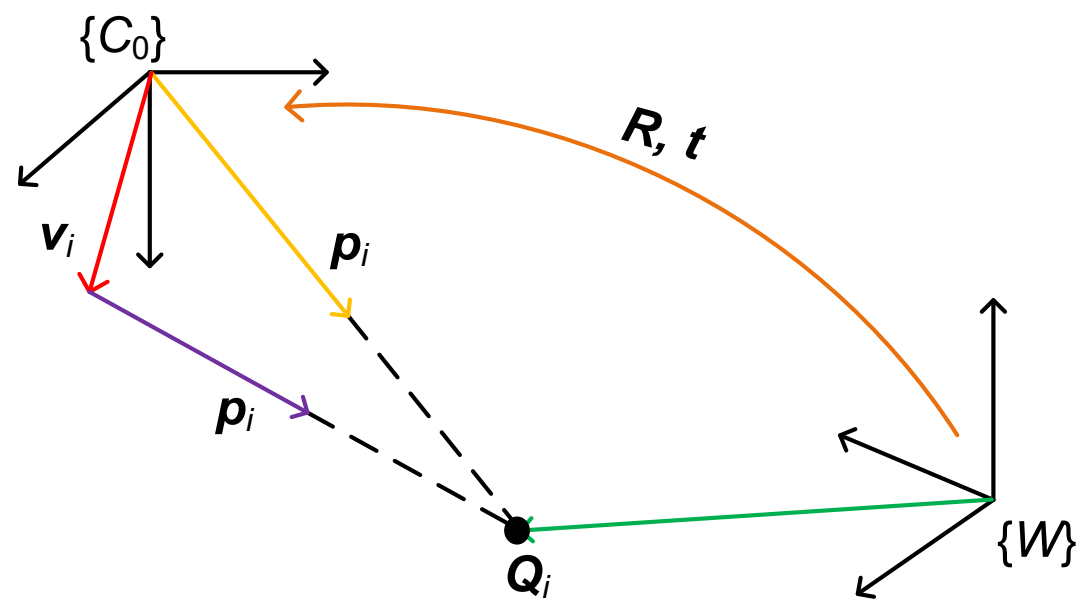

Fig. 2. Representation of image rays in the PnP problem and generalized absolute pose problem.

### A. Notations

Scalars are denoted by lowercase italic letters, vectors by bold lowercase letters, and matrices by bold uppercase letters, with $\boldsymbol{I}$ representing the 3×3 identity matrix. The world frame and the multi-camera reference frame are denoted as $\{W\}$ and $\{C_0\}$, respectively. The subscript $i$ indexes the $i$-th 2-D/3-D point correspondence. VGP denotes the Virtual-point-based Generalized Pose solver, and its subscripts c, q, and r correspond to the Cayley-parameterization-based, quaternion-based, and rotation-matrix-based solutions, respectively.

### B. Single-camera Pose Estimation

Given enough known 3-D points and their 2-D image observations, the orientation and position of a single camera can be obtained by utilizing the projection constraint such that 3-D points $\boldsymbol{Q}_i$ fall on image rays $\boldsymbol{p}_i$ when rotated and translated:

$$\lambda_i \boldsymbol{p}_i = \boldsymbol{R}\boldsymbol{Q}_i + \boldsymbol{t}, \tag{1}$$

where $\boldsymbol{R}$ and $\boldsymbol{t}$ are the camera pose relative to the world frame. $\boldsymbol{R}$ is a 3×3 rotation matrix, and $\boldsymbol{t}$ is a 3×1 translation vector. $\boldsymbol{p}_i$ is the unit (or homogeneous) direction vector of the ray, and $\lambda_i$ is the point depth along the ray.

In the single-camera absolute pose problem or PnP problem, direct minimization methods are popular, which formulate the pose problem as a constrained nonlinear optimization:

$$\min_{\boldsymbol{R},\boldsymbol{t}} \sum_{i=1}^{n} \boldsymbol{r}_i^T(\boldsymbol{R},\mathbf{t})\boldsymbol{r}_i(\boldsymbol{R},\boldsymbol{t}), \tag{2}$$

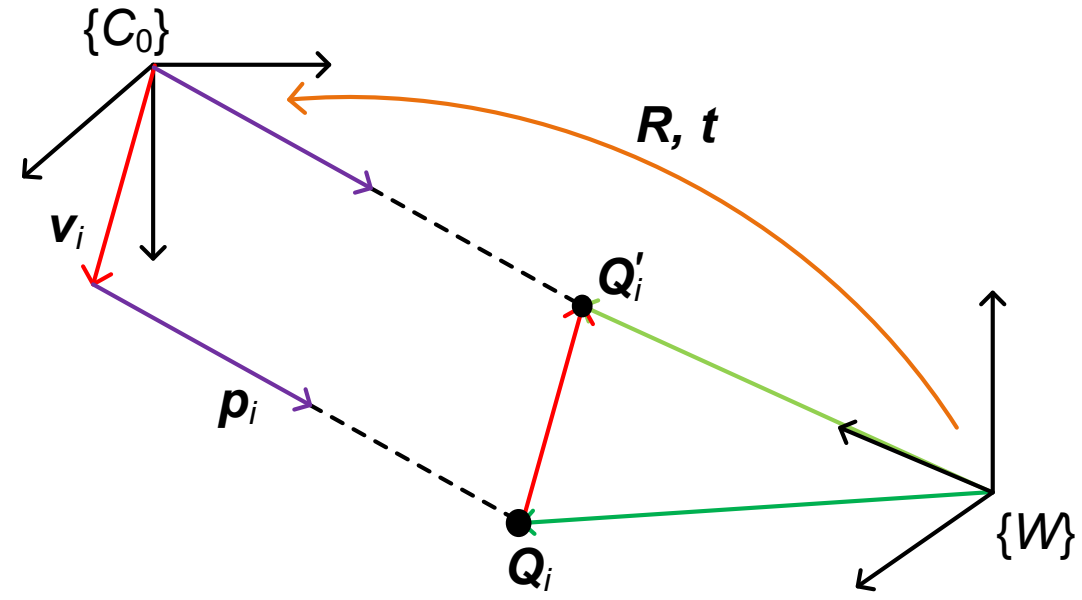

Fig. 3. Illustration of virtual points $\boldsymbol{Q}_i'$. Each $\boldsymbol{Q}_i'$ is obtained by translating the image ray origin against the camera offset direction toward the origin of the reference frame $\{C_0\}$.

where $\boldsymbol{r}_i(\boldsymbol{R},\boldsymbol{t})$ is a residual term involved with the rotation and translation. Its general form is as follows:

$$\boldsymbol{r}_i(\boldsymbol{R},\boldsymbol{t}) = \boldsymbol{X}_i(\boldsymbol{R}\boldsymbol{Q}_i + \boldsymbol{t}), \tag{3}$$

where $\boldsymbol{X}_i$ is a residual matrix, making $\boldsymbol{X}_i(\boldsymbol{R}\boldsymbol{Q}_i + \boldsymbol{t}) = \boldsymbol{0}$ in ideal state.

The cost function $\mathcal{E}^2 = \sum_{i=1}^{n} \boldsymbol{r}_i^T \boldsymbol{r}_i$ is the sum of squared residuals. If we use the vectorized rotation matrix $\boldsymbol{r}_o$, the cost function can be changed into:

$$\mathcal{E}^2 = \sum_{i=1}^{n} \|\boldsymbol{X}_i(\boldsymbol{R}\boldsymbol{Q}_i + \boldsymbol{t})\|^2 = \sum_{i=1}^{n} \|\boldsymbol{X}_i(\boldsymbol{Y}_i\boldsymbol{r}_o + \boldsymbol{t})\|^2, \tag{4}$$

where

$$\boldsymbol{Y}_i = \begin{bmatrix} \boldsymbol{Q}_i^T & \boldsymbol{0} & \boldsymbol{0} \\ \boldsymbol{0} & \boldsymbol{Q}_i^T & \boldsymbol{0} \\ \boldsymbol{0} & \boldsymbol{0} & \boldsymbol{Q}_i^T \end{bmatrix}.$$

Zeroing the derivative of $\mathcal{E}^2$ with respect to $\boldsymbol{t}$, the translation can be represented by the rotation:

$$\boldsymbol{t} = \boldsymbol{T}\boldsymbol{r}_o, \tag{5}$$

where $\boldsymbol{r}_o$ is a 9×1 vector, composed of elements of $\boldsymbol{R}$, and $\boldsymbol{T}$ is a 3×9 transformation matrix, computed from $\boldsymbol{p}_i$ and $\boldsymbol{Q}_i$.

Substituting (5) into (4), the optimization problem reduces to finding the rotation that minimizes $\mathcal{E}^2$. Once the rotation and translation are obtained, the pose problem is solved.

### C. Generalized Pose Estimation

Consider a multi-camera system where all cameras are rigidly mounted on the same carrier platform. As illustrated in Fig. 1, these cameras can be modeled as a generalized camera using rays in space to represent image observations. The projection equation can be extended to:

$$\lambda_i \boldsymbol{p}_i + \boldsymbol{v}_i = \boldsymbol{R}\boldsymbol{Q}_i + \boldsymbol{t}, \tag{6}$$

where $\boldsymbol{v}_i$ is the ray origin (i.e., the offset of the corresponding camera with respect to $\{C_0\}$ shown in Fig. 1). The coordinates of these origins are determined by the extrinsic parameters of the cameras relative to the frame.

Therefore, in the generalized absolute pose problem, image rays originate from multiple projection centers. Consequently, traditional single-camera pose solvers, which assume a single projection center, cannot be directly applied. In a single-camera model, the image ray can be represented solely by a direction vector $\boldsymbol{p}_i$ (yellow). While in a generalized model, the image ray requires a direction vector $\boldsymbol{p}_i$ (purple) and an offset vector $\boldsymbol{v}_i$ (red) to be represented jointly (see Fig. 2).

## III. Virtual-point-based Pose Estimation

To bridge the gap between the generalized and standard projection models, we introduce a virtual point representation supported by the equivalent camera model (Lemma 1). Specifically, we define a virtual point $\boldsymbol{Q}_i' = \boldsymbol{Q}_i - \boldsymbol{R}^{-1}\boldsymbol{v}_i$, which corresponds to shifting the image ray endpoint backwards along the offset (see Fig. 3). Moving $\boldsymbol{v}_i$ to the left side of (6), the generalized projection equation can be rewritten as

$$\lambda_i \boldsymbol{p}_i = \boldsymbol{R}\boldsymbol{Q}_i' + \boldsymbol{t}. \tag{7}$$

### A. Equivalent projection model interpretation

*Lemma 1 (Equivalent projection)*: The virtual point transformation establishes an equivalence between the generalized camera model (6) and the standard pinhole model (1), inducing an equivalent monocular projection that preserves all geometric constraints of the original multi-camera system.

*Proof*: From the generalized projection equation (6), we have $\lambda_i \boldsymbol{p}_i + \boldsymbol{v}_i = \boldsymbol{R}\boldsymbol{Q}_i + \boldsymbol{t}$. Rearranging terms yields $\lambda_i \boldsymbol{p}_i = \boldsymbol{R}(\boldsymbol{Q}_i - \boldsymbol{R}^{-1}\boldsymbol{v}_i) + \boldsymbol{t} = \boldsymbol{R}\boldsymbol{Q}_i' + \boldsymbol{t}$. Equation (7) exhibits the identical mathematical form as the standard pinhole projection (1), where $\boldsymbol{Q}_i'$ serves as the virtual 3-D point observed by an equivalent single camera located at the reference origin $\{C_0\}$. The collinearity constraint between the 3-D point, projection center, and image observation is strictly preserved under this transformation.

*Remark 1 (Model isomorphism)*: The transformation $\mathcal{T}: (\boldsymbol{Q}_i, \boldsymbol{v}_i) \mapsto \boldsymbol{Q}_i'$ induces a geometric isomorphism between the generalized absolute pose problem and the PnP problem. As illustrated in Fig. 3, this mapping effectively consolidates multiple physical projection centers into a unified virtual camera framework while maintaining the ray bundle $\{\boldsymbol{p}_i\}$ invariant. The offset term $\boldsymbol{R}^{-1}\boldsymbol{v}_i$ compensates for the rigid displacement between the $i$-th camera's optical center and the reference frame $\{C_0\}$, thereby centralizing the multi-camera system into a canonical single-camera configuration without altering the underlying projective geometry.

*Remark 2 (Theoretical significance)*: From the perspective of camera modeling, Lemma 1 shows that the distinction between single- and multi-camera systems is merely a matter of point parameterization rather than a geometric essence. The generalized model (6) and the pinhole model (1) share identical algebraic structures; they differ only in whether the 3-D coordinates are expressed in the world frame $\{W\}$ (original) or in a virtual frame adjusted for camera offsets (virtual). This unification shows that any direct minimization method of the form $\min_{\boldsymbol{R},\boldsymbol{t}} \sum_{i=1}^{n} \|\boldsymbol{X}_i(\boldsymbol{R}\boldsymbol{Q}_i + \boldsymbol{t})\|^2$, originally designed for the standard PnP problem, can be extended to generalized camera systems via the virtual point formulation: replace $\boldsymbol{Q}_i$ with $\boldsymbol{Q}_i'$ to obtain $\min_{\boldsymbol{R},\boldsymbol{t}} \sum_{i=1}^{n} \|\boldsymbol{X}_i(\boldsymbol{R}\boldsymbol{Q}_i' + \boldsymbol{t})\|^2$.

*Corollary 1*: The equivalent projection model interpretation rigorously justifies the universality of the proposed framework. Since the transformation preserves the critical geometric constraints that underpin PnP solvers, all optimality conditions and convergence properties of existing single-camera methods are inherently inherited by their generalized counterparts. Based on this established equivalence, we now instantiate the framework with three typical rotation parameterizations—Cayley, quaternion, and rotation matrix—to derive competitive generalized pose solvers.

### B. Cayley-parameterization-based solution (VGPc)

The virtual-point formulation enables uncertainty-aware generalized pose estimation by preserving the covariance structure of the original PnP problem. In standard PnP, DLSU [14] achieves statistical efficiency under heteroscedastic noise via Cayley parameterization; however, its direct extension to generalized cameras is obstructed by the camera offset $\boldsymbol{v}_i$, which potentially introduces anisotropic correlations that violate the independence assumptions underlying the weighted least-squares formulation. By virtue of Lemma 1, the equivalent projection model absorbs the camera offset into the virtual point coordinates via $\boldsymbol{Q}_i' = \boldsymbol{Q}_i - \boldsymbol{R}^{-1}\boldsymbol{v}_i$. This transformation decouples the offset from the residual covariance, reducing the generalized problem to the original PnP structure. Consequently, the statistical optimality of the single-camera solver is rigorously preserved under multi-camera configurations without modification to the weighting scheme. Instantiating this mechanism with DLSU's Cayley-parameterized direct minimization yields VGPc. The detailed implementation proceeds as follows:

The residual $\boldsymbol{r}_i(\boldsymbol{R},\boldsymbol{t})$ in DLSU is expressed as

$$\boldsymbol{r}_i(\boldsymbol{R},\boldsymbol{t}) = \boldsymbol{X}_i\boldsymbol{Y}_i\boldsymbol{r}_o + \boldsymbol{X}_i\boldsymbol{t}, \tag{8}$$

and the covariance of $\boldsymbol{r}_i(\boldsymbol{R},\boldsymbol{t})$ is denoted as $\boldsymbol{\Sigma}_{\boldsymbol{r}_i}$.

Then, the pose problem is formulated into

$$\min_{\boldsymbol{R},\boldsymbol{t}} \sum_{i=1}^{n} \boldsymbol{r}_i^T(\boldsymbol{R},\boldsymbol{t})\boldsymbol{\Sigma}_{\boldsymbol{r}_i}^{-1}\boldsymbol{r}_i(\boldsymbol{R},\boldsymbol{t}). \tag{9}$$

The translation function of DLSU is

$$\boldsymbol{t} = -\boldsymbol{X}^{-1}\boldsymbol{Y}\boldsymbol{r}_o, \tag{10}$$

where $\boldsymbol{X} = \sum_{i=1}^{n} \boldsymbol{X}_i^T\boldsymbol{\Sigma}_{\boldsymbol{r}_i}^{-1}\boldsymbol{X}_i$, $\boldsymbol{Y} = \sum_{i=1}^{n} \boldsymbol{X}_i^T\boldsymbol{\Sigma}_{\boldsymbol{r}_i}^{-1}\boldsymbol{X}_i\boldsymbol{Y}_i$.

Replacing the spatial point $\boldsymbol{Q}_i$ in (8) with virtual point $\boldsymbol{Q}_i'$, we modify the residual as

$$\boldsymbol{r}_i(\boldsymbol{R},\boldsymbol{t}) = \boldsymbol{X}_i\boldsymbol{Y}_i'\boldsymbol{r}_o + \boldsymbol{X}_i\boldsymbol{t}, \tag{11}$$

where

$$\boldsymbol{Y}_i' = \begin{bmatrix} \boldsymbol{Q}_i'^T & \boldsymbol{0} & \boldsymbol{0} \\ \boldsymbol{0} & \boldsymbol{Q}_i'^T & \boldsymbol{0} \\ \boldsymbol{0} & \boldsymbol{0} & \boldsymbol{Q}_i'^T \end{bmatrix}.$$

And replacing $\boldsymbol{Q}_i$ with $\boldsymbol{Q}_i'$ in (10), the translation function in the generalized case can be given by

$$\boldsymbol{t} = -\boldsymbol{X}^{-1}\boldsymbol{Y}'\boldsymbol{r}_o, \tag{12}$$

where $\boldsymbol{Y}' = \sum_{i=1}^{n} \boldsymbol{X}_i^T\boldsymbol{\Sigma}_{\boldsymbol{r}_i}^{-1}\boldsymbol{X}_i\boldsymbol{Y}_i'$.

Eliminating $\boldsymbol{t}$ from $\mathcal{E}^2$, we obtain the cost function that depends only on the rotation:

$$\begin{aligned} \mathcal{E}^2 &= \sum_{i=1}^{n} (\boldsymbol{Y}_i'\boldsymbol{r}_o + \boldsymbol{t})^T\boldsymbol{X}_i^T\boldsymbol{\Sigma}_{\boldsymbol{r}_i}^{-1}\boldsymbol{X}_i(\boldsymbol{Y}_i'\boldsymbol{r}_o + \boldsymbol{t}) \\ &= \sum_{i=1}^{n} (\boldsymbol{Y}_i'\boldsymbol{r}_o - \boldsymbol{X}^{-1}\boldsymbol{Y}'\boldsymbol{r}_o)^T\boldsymbol{X}_i^T\boldsymbol{\Sigma}_{\boldsymbol{r}_i}^{-1}\boldsymbol{X}_i(\boldsymbol{Y}_i'\boldsymbol{r}_o - \boldsymbol{X}^{-1}\boldsymbol{Y}'\boldsymbol{r}_o), \end{aligned} \tag{13}$$

where $\boldsymbol{Y}_i'\boldsymbol{r}_o = \boldsymbol{Y}_i\boldsymbol{r}_o - \boldsymbol{v}_i$, $\boldsymbol{X}^{-1}\boldsymbol{Y}'\boldsymbol{r}_o = \boldsymbol{X}^{-1}\boldsymbol{Y}\boldsymbol{r}_o - \boldsymbol{X}^{-1}\boldsymbol{V}$, $\boldsymbol{V} = \sum_{i=1}^{n} \boldsymbol{X}_i^T\boldsymbol{\Sigma}_{\boldsymbol{r}_i}^{-1}\boldsymbol{X}_i\boldsymbol{v}_i$.

Following DLSU, we use the Cayley parameterization and multiply the cost function by $(1+||\boldsymbol{s}||^2)^2$, where $\boldsymbol{s} = [s_1, s_2, s_3]^T$ is a 3×1 vector of Cayley rotation parameters.

Hence, the generalized pose problem becomes a minimization problem of $\boldsymbol{s}$:

$$\min_{\boldsymbol{s}} \boldsymbol{r}_s^T \boldsymbol{M} \boldsymbol{r}_s, \tag{14}$$

where $\boldsymbol{r}_s = [1, s_1, s_2, s_3, s_1^2, s_1 s_2, s_1 s_3, s_2^2, s_2 s_3, s_3^2]^T$, $\boldsymbol{M}$ is a coefficient matrix, computed from $\boldsymbol{p}_i$, $\boldsymbol{v}_i$ and $\boldsymbol{Q}_i$.

We can constrain the derivative of $\mathcal{E}^2 = \boldsymbol{r}_s^T \boldsymbol{M} \boldsymbol{r}_s$ to be zero and obtain three third-order polynomials:

$$\left(\frac{\partial \boldsymbol{r}_s}{\partial \boldsymbol{s}}\right) \boldsymbol{M} \boldsymbol{r}_s = \boldsymbol{0}_{3\times 1}, \tag{15}$$

which can be solved by the solver generated by [12].

### C. Quaternion-based solution (VGPq)

Global optimality in PnP solvers is typically attained under the first-order optimality conditions of rotation parameters. OPnP [8] achieves this property using non-unit quaternion parameterization and serves as a widely adopted baseline. Extending such certified optimality to generalized cameras is nontrivial: the camera offset $\boldsymbol{v}_i$ in (6) induces spatially varying coupling between translation and rotation across multiple projection centers, disrupting the problem structure required for global optimality certificates. Through the equivalent projection model (7), we show that expressing the generalized problem via virtual points $\boldsymbol{Q}_i'$ restores the identical algebraic form as the standard PnP. The offset compensation term $\boldsymbol{R}^{-1}\boldsymbol{v}_i$ constitutes a rigid-body displacement that introduces no additional local minima in the rotation space; consequently, the global optimality guarantee is rigorously inherited by the generalized formulation. Instantiating this mechanism with OPnP's quaternion-based direct minimization yields VGPq. The detailed implementation proceeds as follows:

In OPnP, a non-unit quaternion $(q_1, q_2, q_3, q_4)$ is used to parameterize the rotation, and the pose problem is formulated into a minimization problem:

$$\begin{aligned} \min_{q_1,q_2,q_3,q_4,\hat{t}_x,\hat{t}_y} & \sum_{i=1}^{n} \left[\left(1 + \bar{\boldsymbol{r}}_3^T \widetilde{\boldsymbol{Q}}_i\right) u_i - \bar{\boldsymbol{r}}_1^T \boldsymbol{Q}_i - \hat{t}_x\right]^2 \\ & + \sum_{i=1}^{n} \left[\left(1 + \bar{\boldsymbol{r}}_3^T \widetilde{\boldsymbol{Q}}_i\right) v_i - \bar{\boldsymbol{r}}_2^T \boldsymbol{Q}_i - \hat{t}_y\right]^2, \end{aligned} \tag{16}$$

where $\bar{\boldsymbol{r}}_j = s\boldsymbol{r}_j, j = 1,2,3$, $\boldsymbol{r}_j$ is the $j$-th row of $\boldsymbol{R}$, $s = q_1^2 + q_2^2 + q_3^2 + q_4^2$, $\widetilde{\boldsymbol{Q}}_i = \boldsymbol{Q}_i - \overline{\boldsymbol{Q}}$, $\overline{\boldsymbol{Q}} = \frac{1}{n}\sum_{i=1}^{n} \boldsymbol{Q}_i$, $[u_i, v_i]^T$ is the normalized coordinates of $\boldsymbol{Q}_i$, $\hat{t}_x = st_x$, $\hat{t}_y = st_y$, and $t_x$, $t_y$ are the $\boldsymbol{x}-$ and $\boldsymbol{y}-$ components of the translation vector.

In (16), items $\hat{t}_x, \hat{t}_y$ related with the translation can be represented by items $\bar{\boldsymbol{r}}_j, j = 1,2,3$ related to the rotation:

$$\begin{aligned} \hat{t}_x &= \bar{u} + \bar{\boldsymbol{r}}_3^T \left(\frac{1}{n}\sum_{i=1}^{n} u_i \widetilde{\boldsymbol{Q}}_i\right) - \bar{\boldsymbol{r}}_1^T \overline{\boldsymbol{Q}}, \\ \hat{t}_y &= \bar{v} + \bar{\boldsymbol{r}}_3^T \left(\frac{1}{n}\sum_{i=1}^{n} v_i \widetilde{\boldsymbol{Q}}_i\right) - \bar{\boldsymbol{r}}_2^T \overline{\boldsymbol{Q}}, \end{aligned} \tag{17}$$

where $\bar{u} = \frac{1}{n}\sum_{i=1}^{n} u_i$, $\bar{v} = \frac{1}{n}\sum_{i=1}^{n} v_i$.

Replacing the spatial point $\boldsymbol{Q}_i$ in (16) with virtual point $\boldsymbol{Q}_i'$, we modify the cost functions as

$$\begin{aligned} & \sum_{i=1}^{n} \left[\left(1 + \bar{\boldsymbol{r}}_3^T \widetilde{\boldsymbol{Q}}_i'\right) u_i - \bar{\boldsymbol{r}}_1^T \boldsymbol{Q}_i' - \hat{t}_x\right]^2 \\ & + \sum_{i=1}^{n} \left[\left(1 + \bar{\boldsymbol{r}}_3^T \widetilde{\boldsymbol{Q}}_i\right) v_i - \bar{\boldsymbol{r}}_2^T \boldsymbol{Q}_i' - \hat{t}_y\right]^2. \end{aligned} \tag{18}$$

Replacing $\boldsymbol{Q}_i$ with $\boldsymbol{Q}_i'$ in (17) and using orthogonal constraints that $\bar{\boldsymbol{r}}_i^T \bar{\boldsymbol{r}}_j = 0, \bar{\boldsymbol{r}}_i^T \bar{\boldsymbol{r}}_i = s^2$, the translation items $\hat{t}_x, \hat{t}_y$ in the generalized case can be given by

$$\begin{aligned} \hat{t}_x &= \bar{u} + \bar{\boldsymbol{r}}_3^T \left(\frac{1}{n}\sum_{i=1}^{n} u_i \widetilde{\boldsymbol{Q}}_i'\right) - \bar{\boldsymbol{r}}_1^T \overline{\boldsymbol{Q}}' \\ &= \bar{u} + \bar{\boldsymbol{r}}_3^T \left(\frac{1}{n}\sum_{i=1}^{n} u_i \left(\boldsymbol{Q}_i' - \overline{\boldsymbol{Q}}'\right)\right) - \bar{\boldsymbol{r}}_1^T \overline{\boldsymbol{Q}}' \\ &= \bar{u} + \bar{\boldsymbol{r}}_3^T \left(\frac{1}{n}\sum_{i=1}^{n} u_i \left(\widetilde{\boldsymbol{Q}}_i - \boldsymbol{R}^{-1}(\boldsymbol{v}_i - \bar{\boldsymbol{v}})\right)\right) \\ &\quad - \bar{\boldsymbol{r}}_1^T (\overline{\boldsymbol{Q}} - \boldsymbol{R}^{-1}\bar{\boldsymbol{v}}) \\ &= \bar{u} + \bar{\boldsymbol{r}}_3^T \left(\frac{1}{n}\sum_{i=1}^{n} u_i \widetilde{\boldsymbol{Q}}_i\right) - s\left(\frac{1}{n}\sum_{i=1}^{n} u_i \widetilde{\boldsymbol{v}}_i\right)_3 \\ &\quad - \bar{\boldsymbol{r}}_1^T \overline{\boldsymbol{Q}} + s(\bar{\boldsymbol{v}})_1, \end{aligned} \tag{19}$$

$$\begin{aligned} \hat{t}_y &= \bar{v} + \bar{\boldsymbol{r}}_3^T \left(\frac{1}{n}\sum_{i=1}^{n} v_i \widetilde{\boldsymbol{Q}}_i'\right) - \bar{\boldsymbol{r}}_2^T \overline{\boldsymbol{Q}}' \\ &= \bar{v} + \bar{\boldsymbol{r}}_3^T \left(\frac{1}{n}\sum_{i=1}^{n} v_i \widetilde{\boldsymbol{Q}}_i\right) - s\left(\frac{1}{n}\sum_{i=1}^{n} v_i \widetilde{\boldsymbol{v}}_i\right)_3 \\ &\quad - \bar{\boldsymbol{r}}_2^T \overline{\boldsymbol{Q}} + s(\bar{\boldsymbol{v}})_2, \end{aligned} \tag{20}$$

where $\widetilde{\boldsymbol{v}}_i = \boldsymbol{v}_i - \bar{\boldsymbol{v}}$, $\bar{\boldsymbol{v}} = \frac{1}{n}\sum_{i=1}^{n} \boldsymbol{v}_i$, $(\cdot)_k$ means the $k$-th element of the vector inside the parentheses.

Substituting (19) and (20) into (18), the optimisation problem is left only about the rotation:

$$\min_{q_1,q_2,q_3,q_4} \boldsymbol{r}_q^T \boldsymbol{M} \boldsymbol{r}_q, \tag{21}$$

where $\boldsymbol{r}_q = [1, q_1^2, q_1 q_2, q_1 q_3, q_1 q_4, q_2^2, q_2 q_3, q_2 q_4, q_3^2, q_3 q_4, q_4^2]^T$, $\boldsymbol{M}$ is the coefficient matrix, computed from $\boldsymbol{p}_i$, $\boldsymbol{v}_i$ and $\boldsymbol{Q}_i$.

Taking the partial derivatives of $\mathcal{E}^2 = \boldsymbol{r}_q^T \boldsymbol{M} \boldsymbol{r}_q$ with respect to $(q_1, q_2, q_3, q_4)$ yields four third-order polynomial equations:

$$\frac{\partial \mathcal{E}^2}{\partial q_1} = 0, \frac{\partial \mathcal{E}^2}{\partial q_2} = 0, \frac{\partial \mathcal{E}^2}{\partial q_3} = 0, \frac{\partial \mathcal{E}^2}{\partial q_4} = 0. \tag{22}$$

We can solve this system under the first-order optimality condition to obtain the camera pose.

### D. Rotation-matrix-based solution (VGPr)

Real-time pose estimation via sequential quadratic programming relies on the homogeneous quadratic structure of the standard PnP problem. SQPnP [13], for instance, operates directly on the rotation matrix and achieves efficient convergence under this structure, with official support in OpenCV. Extending such efficiency to generalized cameras is obstructed by the camera offset $\boldsymbol{v}_i$ in (6), which introduces additional linear terms into the quadratic programming formulation and transforms the original homogeneous problem

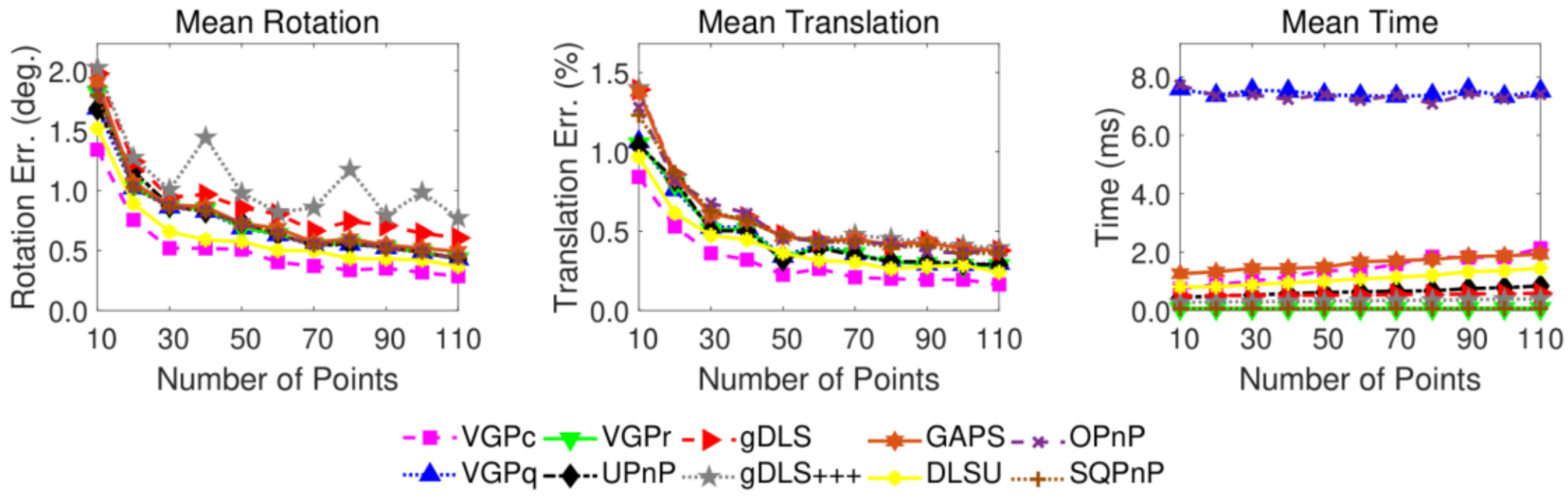

Fig. 4. Pose estimation results on synthetic data with varying numbers of points. The running time per estimation is averaged over 100 independent trials.

into an inhomogeneous one. This structural perturbation complicates the SQP iterations and degrades real-time performance. By exploiting the virtual point mapping, we eliminate the inhomogeneous terms by encoding the multi-camera geometry into $\boldsymbol{Q}_i' = \boldsymbol{Q}_i - \boldsymbol{R}^{-1}\boldsymbol{v}_i$, thereby restoring the standard homogeneous quadratic form. Consequently, the computational complexity of the original SQPnP solver is rigorously preserved under generalized configurations, with camera offsets absorbed into a precomputed coefficient matrix that does not perturb the SQP solution process. Instantiating this mechanism with SQPnP's rotation-matrix-based sequential quadratic programming yields VGPr. The detailed generalized formulation proceeds as follows:

The optimization problem in SQPnP is given by

$$\min_{\boldsymbol{r}_o,\boldsymbol{t}} \sum_{i=1}^{n} \|\boldsymbol{X}_i(\boldsymbol{Y}_i\boldsymbol{r}_o + \boldsymbol{t})\|^2 = \sum_{i=1}^{n} (\boldsymbol{Y}_i\boldsymbol{r}_o + \boldsymbol{t})^T \boldsymbol{G}_i (\boldsymbol{Y}_i\boldsymbol{r}_o + \boldsymbol{t}), \quad (23)$$

where $\boldsymbol{G}_i = \boldsymbol{X}_i^T\boldsymbol{X}_i$.

The translation function is written as

$$\boldsymbol{t} = \boldsymbol{T}_r\boldsymbol{r}_o, \quad (24)$$

where $\boldsymbol{T}_r = -(\sum_{i=1}^{n}\boldsymbol{G}_i)^{-1}(\sum_{i=1}^{n}\boldsymbol{G}_i\boldsymbol{Y}_i)$.

Replacing the spatial point $\boldsymbol{Q}_i$ in (23) with virtual point $\boldsymbol{Q}_i'$, the cost function in the generalized case can be rewritten as

$$\begin{aligned} \mathcal{E}^2 &= \sum_{i=1}^{n} \|\boldsymbol{X}_i(\boldsymbol{Y}_i'\boldsymbol{r}_o + \boldsymbol{t})\|^2 \\ &= \sum_{i=1}^{n} (\boldsymbol{Y}_i\boldsymbol{r}_o + \boldsymbol{Z}_i\boldsymbol{t}')^T \boldsymbol{G}_i (\boldsymbol{Y}_i\boldsymbol{r}_o + \boldsymbol{Z}_i\boldsymbol{t}'), \end{aligned} \quad (25)$$

where $\boldsymbol{Z}_i = [\boldsymbol{I}, -v_i]$, $\boldsymbol{t}' = [\boldsymbol{t}, 1]^T$ is the homogeneous form of the translation vector.

Zeroing the derivative of $\mathcal{E}^2$ with respect to $\boldsymbol{t}'$ yields

$$\begin{aligned} &\sum_{i=1}^{n} \boldsymbol{Z}_i^T\boldsymbol{G}_i(\boldsymbol{Y}_i\boldsymbol{r}_o + \boldsymbol{Z}_i\,\boldsymbol{t}') = \boldsymbol{0} \\ \Leftrightarrow &\left(\sum_{i=1}^{n} \boldsymbol{Z}_i^T\boldsymbol{G}_i\boldsymbol{Z}_i\right)\boldsymbol{t}' = -\left(\sum_{i=1}^{n} \boldsymbol{Z}_i^T\boldsymbol{G}_i\boldsymbol{Y}_i\right)\boldsymbol{r}_o. \end{aligned} \quad (26)$$

The translation can be expressed in terms of the rotation vector:

$$\boldsymbol{t}' = \boldsymbol{T}_r'\boldsymbol{r}_o, \quad (27)$$

where $\boldsymbol{T}_r' = -(\sum_{i=1}^{n}\boldsymbol{Z}_i^T\boldsymbol{G}_i\boldsymbol{Z}_i)^{-1}(\sum_{i=1}^{n}\boldsymbol{Z}_i^T\boldsymbol{G}_i\boldsymbol{Y}_i)$.

Substituting (27) into (25) yields an optimization problem about the rotation only:

$$\min_{\boldsymbol{r}_o} \boldsymbol{r}_o^T\boldsymbol{M}\boldsymbol{r}_o, \quad (28)$$

where $\boldsymbol{M} = \sum_{i=1}^{n}(\boldsymbol{Y}_i + \boldsymbol{Z}_i\boldsymbol{T}')^T\boldsymbol{G}_i(\boldsymbol{Y}_i + \boldsymbol{Z}_i\boldsymbol{T}')$.

We first compute initial values of $\boldsymbol{r}_o$ via SVD($\boldsymbol{M}$) and refine them to global optimality via SQP. Although $\boldsymbol{t}$ can be recovered by constraining the fourth element of $\boldsymbol{t}'$ to 1 once $\boldsymbol{r}_o$ is determined, this normalization compromises precision. To attain the optimal solution, we rearrange (26) to yield:

$$\boldsymbol{t} = \boldsymbol{T}_r\boldsymbol{r}_o + \boldsymbol{t}_v \quad (29)$$

where $\boldsymbol{t}_v = (\sum_{i=1}^{n}\boldsymbol{G}_i)^{-1}(\sum_{i=1}^{n}\boldsymbol{G}_i\boldsymbol{v}_i)$.

Moreover, as discussed in [20], there is a case in which the scale in the camera frame is not consistent with the world frame, yielding an unknown scale and forming a generalized pose $(\boldsymbol{R}, \boldsymbol{t})$ and scale $(s)$ problem. In this case, the projection equation is written as

$$\lambda_i\boldsymbol{p}_i + s\boldsymbol{v}_i = \boldsymbol{R}\boldsymbol{Q}_i + \boldsymbol{t}. \quad (30)$$

Letting $\boldsymbol{Q}_i' = \boldsymbol{Q}_i - \boldsymbol{R}^{-1}(s\boldsymbol{v}_i)$, the $\boldsymbol{t}'$ in (25) becomes $\boldsymbol{t}' = [\boldsymbol{t}, s]^T$, which can be directly obtained from (27), thus enabling the solution of the generalized pose and scale problem.

## IV. Experiments

In this section, comprehensive experiments are performed on synthetic data and real-world scenarios to validate the performance of the proposed solvers.

### *A. Experiments with Synthetic Data*

In this section, we present experimental results on synthetic data. The experiments were implemented in MATLAB on a computer equipped with an Intel Core Ultra 9 275HX CPU @ 2.70 GHz. The simulated camera's parameters include a 640×480-pixel image size and an 800-pixel focal length. The 3-D points are uniformly distributed within the region $[-2,2]m \times [-2,2]m \times [-2,2]m$. The cameras (i.e., ray origins) are randomly placed within the volume $[-1,1]m \times [-1,1]m \times [-1,1]m$. 2-D/3-D correspondences are generated by projecting the 3-D points with random rotations and translations onto randomly selected cameras. To assess estimation accuracy, the relative rotation error $\varepsilon_{rot}(\circ)$ was computed as $\mathrm{acos}\big((trace(\boldsymbol{R}^T\boldsymbol{R}_{true}) - 1)/2\big)$. Additionally, the relative position error $\varepsilon_{tra}(\%)$ was measured as $\|\boldsymbol{t} - \boldsymbol{t}_{true}\|/\|\boldsymbol{t}\| \times 100$. The results are averaged over 100 trials. The following methods are compared:

1) *PnP solvers*: DLSU [14], OPnP [8], and SQPnP [13],

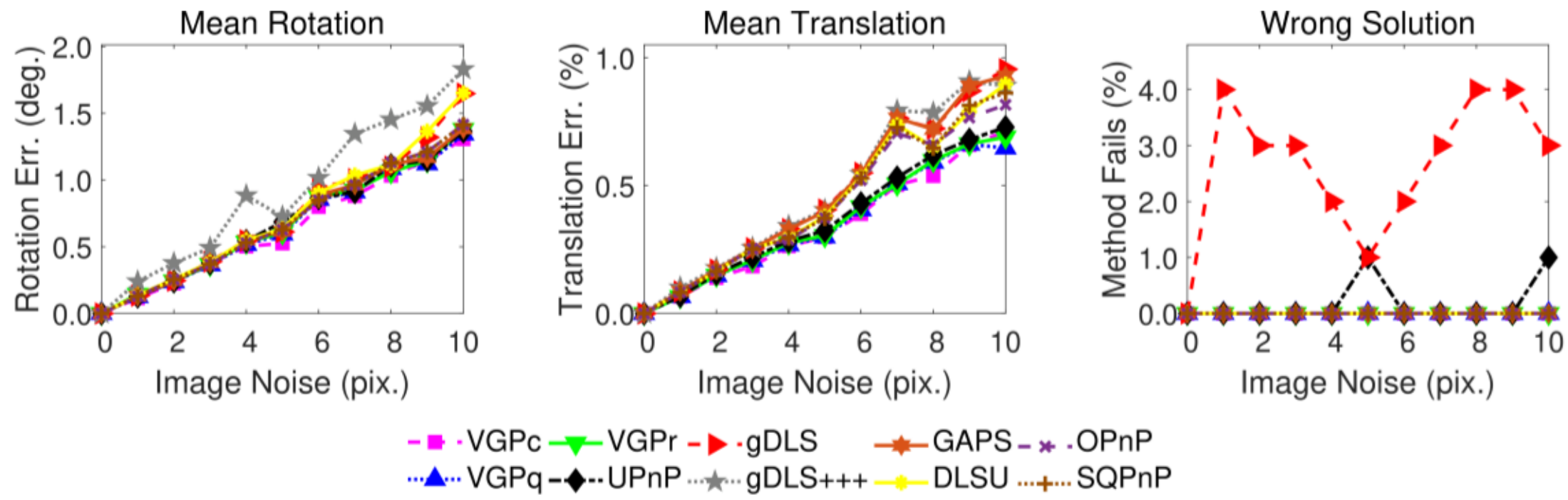

Fig. 5. Pose estimation results on synthetic data with increasing Gaussian image noise. A solution is classified as incorrect (or a failure) when the rotation error exceeds 10°.

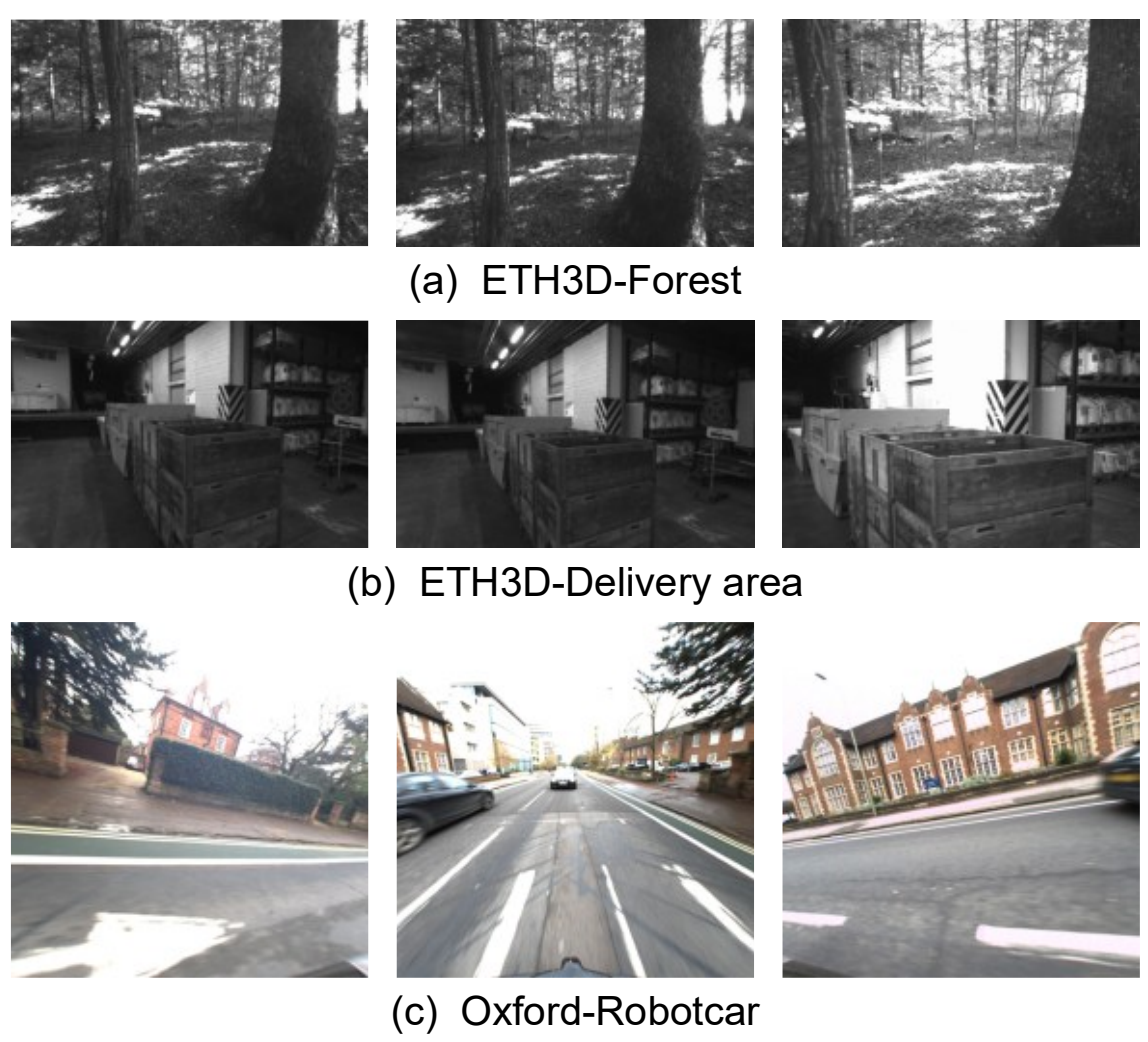

Fig. 6. Typical images of the used datasets are presented.

which all address single-camera pose problems.

2) *Generalized pose solvers*: UPnP [19], gDLS [20], gDLS+++ [22], GAPS [23], and the proposed VGPc, VGPq, and VGPr, which all address generalized pose problems.

To comprehensively evaluate the performance of the proposed VGP solvers, we conduct systematic experiments on synthetic data with controlled parameters. The synthetic experiments are designed to assess three critical aspects: (1) estimation accuracy and computational efficiency with varying numbers of point correspondences and (2) robustness against increasing levels of image noise. The following subsections detail the comparative results against state-of-the-art PnP solvers (DLSU, OPnP, SQPnP) and generalized pose solvers (UPnP, gDLS, gDLS+++, GAPS).

*Simulations with varying point numbers*: We vary the number of point correspondences from $n = 10$ to 110 and introduce heteroscedastic noise by partitioning the 2-D projections into 10 equally sized subsets with varying noise levels, from $\sigma = 0$ to 10 pixels. The rotation and translation estimation results, along with the computation time, are presented in Fig. 4. The results demonstrate that the proposed VGP methods inherit the superior accuracy and computational efficiency from their original single-camera methods (DLSU, OPnP, and SQPnP), while outperforming existing generalized pose estimation approaches. Specifically, VGPc achieves the highest accuracy with a rotation error of only 0.29° and a translation error of 0.16% at $n = 110$, representing improvements of 33% and 43%, respectively, over the state-of-the-art generalized solver UPnP (0.43° and 0.29%). VGPq and VGPr maintain comparable accuracy, while VGPr emerges as the optimal real-time solution, operating at only 0.08 ms at $n = 110$, 5× faster than gDLS+++ (0.40 ms) and 12× faster than UPnP (0.84 ms) under the same conditions, without accuracy degradation.

*Simulations with increasing image noise*: We fix the point number $n = 20$ and add isotropic noise to the 2-D projections by increasing the standard deviation from $\sigma = 0$ to 10 pixels. The simulation results are shown in Fig. 5. The results demonstrate that the proposed VGP solvers maintain superior robustness to increasing image noise, consistently achieving lower estimation errors across all noise levels. Specifically, VGPc achieves the highest accuracy and best noise resilience among all tested solvers, maintaining stable estimates with zero failures. All VGP solvers exhibit exceptional numerical stability throughout the entire noise range, whereas gDLS+++, gDLS, and even UPnP suffer from varying degrees of instability, reaching failure rates of 53%, 4%, and 1% at $\sigma = 10$ pixels, respectively. These results validate that the proposed framework effectively preserves the robustness of the original single-camera solvers while extending their applicability to generalized camera systems.

### B. Experiments in Real-world Scenarios

We further validate the proposed solvers on real-world scenarios using public datasets and an actual multi-camera system, assessing their real-world performance.

*Validations with public datasets*: To evaluate the proposed solvers in real-world scenarios, we test them on publicly available datasets featuring known 3-D structures, detected 2-D image observations, and reference camera poses. Specifically, experiments are conducted on the ETH3D Many-view datasets [26] and the Oxford RobotCar dataset [27]. The former provides image data captured by a four-camera system across ten diverse indoor and outdoor scenes (e.g., delivery areas, forests). The latter comprises sequential images from three synchronized cameras (rear-left, rear, and rear-right) mounted on a vehicle traversing Oxford, UK, acquired over multiple traversals under varying seasonal and illumination conditions. Fig. 6 presents typical images from these datasets. We conduct tests by randomly sampling 2-D/3-D correspondences ($n = 10$ to $110$) from selected sequences of the ETH3D and Oxford datasets. The test results are presented in Figs. 7 and 8, respectively. We select two representative scenes from the ETH3D datasets (an indoor delivery area and an outdoor forest environment) and compare the estimation accuracy of VGPc, VGPq, and VGPr with that of existing solvers. As shown in Fig. 7, the proposed VGP solvers consistently achieve competitive or superior estimation accuracy compared to other solvers. Specifically, VGPc achieves the highest accuracy in both scenes, with rotation errors of only 0.06° (indoor) and 0.07° (outdoor) at $n = 110$. Notably, VGPq and VGPr maintain accuracy comparable to GAPS [23] and UPnP [19] in the indoor scenario, while slightly outperforming the latter in the outdoor forest scene. In contrast, gDLS+++ [22] exhibits severe instability in both real-world conditions, producing highly erratic estimates with pose errors that fluctuate significantly across different point numbers. Additionally, tests on the Oxford dataset validate the generalization capability of the proposed solvers to dynamic driving scenarios with vehicle motion. As shown in Fig. 8, VGPc consistently achieves the highest accuracy, with rotation and translation errors of merely 0.07° and 0.05% at $n = 110$, respectively, and outperforms GAPS by comparable margins. VGPq and VGPr maintain accuracy slightly better than UPnP and GAPS. Furthermore, aggregating results from all three datasets, Fig. 9 presents the quantitative relative error reduction with respect to UPnP at $n = 110$. As illustrated in Fig. 9, VGPc achieves substantial improvements of 23% in rotation and 57% in translation accuracy, respectively. Notably, VGPr delivers positive gains alongside superior computational efficiency. These results confirm the validity of the proposed framework.

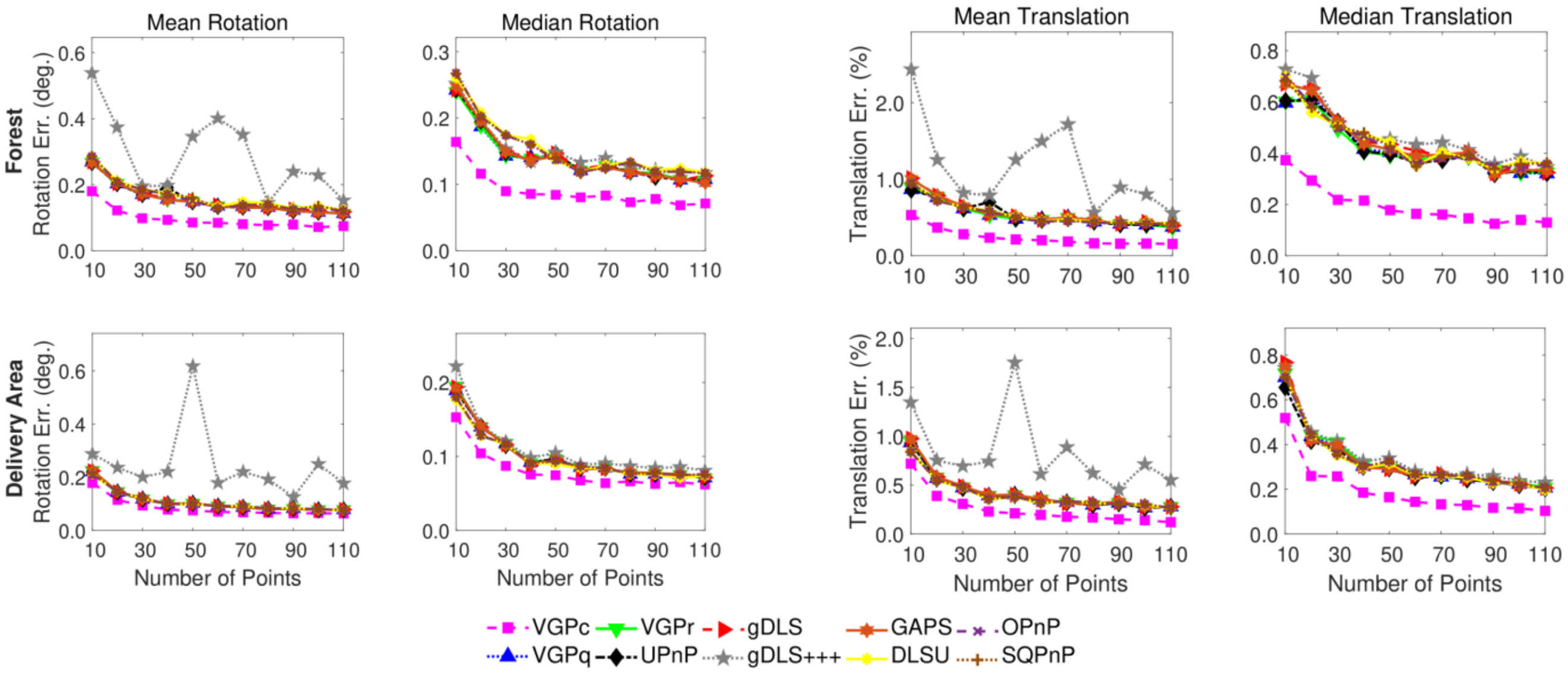

Fig. 7. Pose estimation results in the ETH3D Many-view datasets with varying point numbers. The top and bottom rows show results for the outdoor (Forest) and indoor (Delivery area) sequences, respectively.

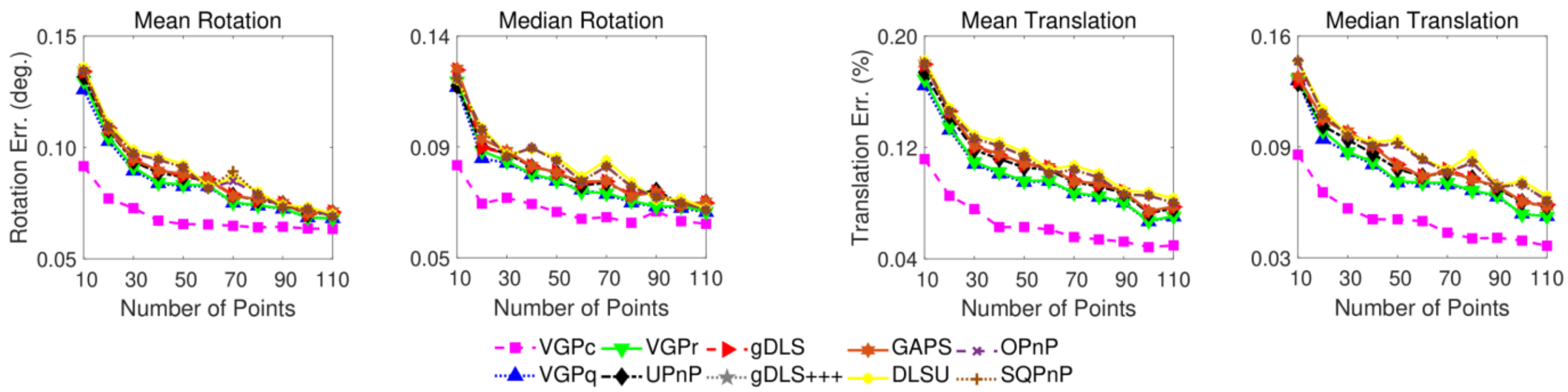

Fig. 8. Pose estimation results in the Oxford RobotCar dataset with varying point numbers.

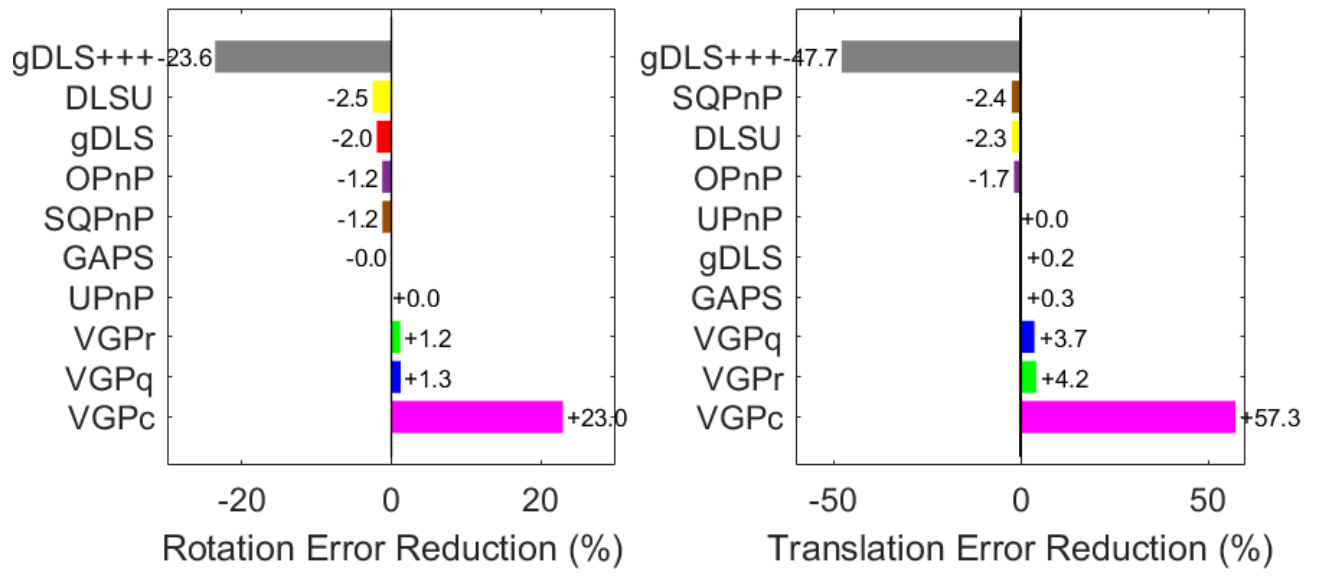

Fig. 9. Relative performance improvement over UPnP baseline at $n = 110$. Positive values indicate error reduction averaged over ETH3D and Oxford datasets.

*Validations with the DMAIS*: High-precision pose estimation traditionally requires balancing spatial resolution against field of view (FoV), an inherent trade-off in single-camera systems. To circumvent this limitation, we developed a divergent multi-aperture imaging system (DMAIS) [25] comprising five long-focal-length cameras (150mm) with narrow fields of view ($5.4° \times 4.0°$). As shown in Fig. 10, each camera is oriented in a dispersed direction such that its individual, non-overlapping observations collectively span a wide effective FoV ($90° \times 90°$). This configuration enables high-resolution imaging across a broad angular range without the optical aberrations typical of wide-angle lenses. We mounted the DMAIS on a quadrotor UAV that was hovering at approximately 350m to observe ground targets (Fig. 11). For ground-truth reference, we installed a high-precision position and orientation system (POS) aboard the UAV to record its position. On the ground, we deployed diagonal markers (position-surveyed via GNSS-RTK) as artificial landmarks for visual localization. Position accuracy is quantified as the Euclidean distance $\varepsilon_{pos}$ between the estimated pose and POS measurements. As no high-precision orientation reference was available, we infer rotational error through angular deviation $\varepsilon_{ang} = \varepsilon_{rep} \times \theta_{res}$, where $\varepsilon_{rep}$ denotes the reprojection error, and $\theta_{res}$ represents the angular resolution computed from camera intrinsics. We performed independent flight trials, varying the active camera subsets. For example, configuration *#012* utilized only cameras $C_0$, $C_1$, and $C_2$. The results are summarized in Table I, with the smallest values highlighted in bold and the second smallest underlined. As the camera configuration scales from the monocular setup (#0) to the multi-camera arrays (#123 and #01234), the estimation accuracy improves for all methods due to increased geometric redundancy, with position errors decreasing from sub-meter to centimeter levels. Notably, under the large field-of-view configurations (#123 and #01234), UPnP, VGPc, VGPq, and VGPr exhibit nearly identical accuracy levels, with mirror differences. This phenomenon arises because the DMAIS employs long-focal-length, narrow-field-of-view cameras, for which the image noise is approximately isotropic, consistent with the synthetic experimental results shown in Fig. 5. Under such conditions, the weighted least-squares formulation of VGPc reduces to ordinary least squares, neutralizing its theoretical advantage in uncertainty modeling. Consequently, VGPr achieves comparable accuracy to UPnP while demonstrating superior computational efficiency. Meanwhile, conventional single-camera solvers (DLSU, OPnP, SQPnP) become inapplicable in multi-camera setups, and other generalized methods such as gDLS and GAPS suffer from accuracy degradation or numerical instability. These findings validate that the virtual-point formulation effectively inherits the superior characteristics of single-camera PnP algorithms while seamlessly scaling to multi-camera topologies, providing a reliable solution for multi-camera applications.

TABLE I
POSE ESTIMATION RESULTS WITH DIFFERENT CAMERA CONFIGURATIONS

| METHODS | | DLSU [14] | OPnP [8] | SQPnP [13] | UPnP [19] | gDLS [20] | gDLS+++ [22] | GAPS [23] | VGPc | VGPq | VGPr |
|---|---|---|---|---|---|---|---|---|---|---|---|
| *#0* | $\varepsilon_{pos}$(m) | 0.363 | **0.183** | 0.184 | 0.361 | - | - | 0.311 | 0.363 | **0.183** | 0.184 |
| | $\varepsilon_{ang}(')$ | 3.860 | **3.203** | 3.299 | 6.201 | - | - | 7.270 | 3.842 | **3.203** | 3.299 |
| | $\varepsilon_{rep}$(px) | 49.220 | **40.836** | 42.060 | 79.073 | - | - | 92.698 | 48.984 | **40.836** | 42.060 |
| | $time(ms)$ | 0.902 | 7.632 | **0.073** | 0.491 | - | - | 1.336 | 0.931 | 7.852 | **0.073** |
| *#123* | $\varepsilon_{pos}$(m) | - | - | - | **0.027** | 0.349 | - | 0.084 | **0.027** | **0.027** | **0.027** |
| | $\varepsilon_{ang}(')$ | - | - | - | **0.270** | 2.751 | - | 0.569 | 0.271 | 0.271 | 0.271 |
| | $\varepsilon_{rep}$(px) | - | - | - | **3.448** | 35.078 | - | 7.257 | 3.454 | 3.449 | 3.451 |
| | $time(ms)$ | - | - | - | 0.641 | 0.547 | - | 1.496 | 1.295 | 7.773 | **0.077** |
| *#01234* | $\varepsilon_{pos}$(m) | - | - | - | 0.047 | 0.055 | 0.062 | 0.056 | 0.047 | **0.046** | **0.046** |
| | $\varepsilon_{ang}(')$ | - | - | - | **0.251** | 0.280 | 0.635 | 0.281 | **0.251** | **0.251** | **0.251** |
| | $\varepsilon_{rep}$(px) | - | - | - | **3.201** | 3.573 | 8.091 | 3.585 | **3.201** | **3.201** | **3.201** |
| | $time(ms)$ | - | - | - | 0.705 | 0.562 | 0.360 | 2.068 | 1.545 | 8.066 | **0.078** |

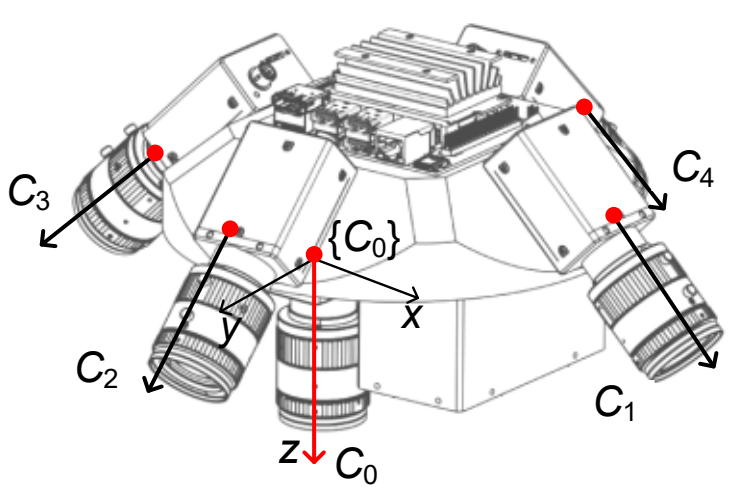

Fig. 10. Structure diagram of DMAIS. The central camera, denoted as $C_0$, observes the ground vertically downward. The four surrounding cameras, labeled $C_1$, $C_2$, $C_3$, and $C_4$, are positioned to face forward, backward, left, and right, respectively.

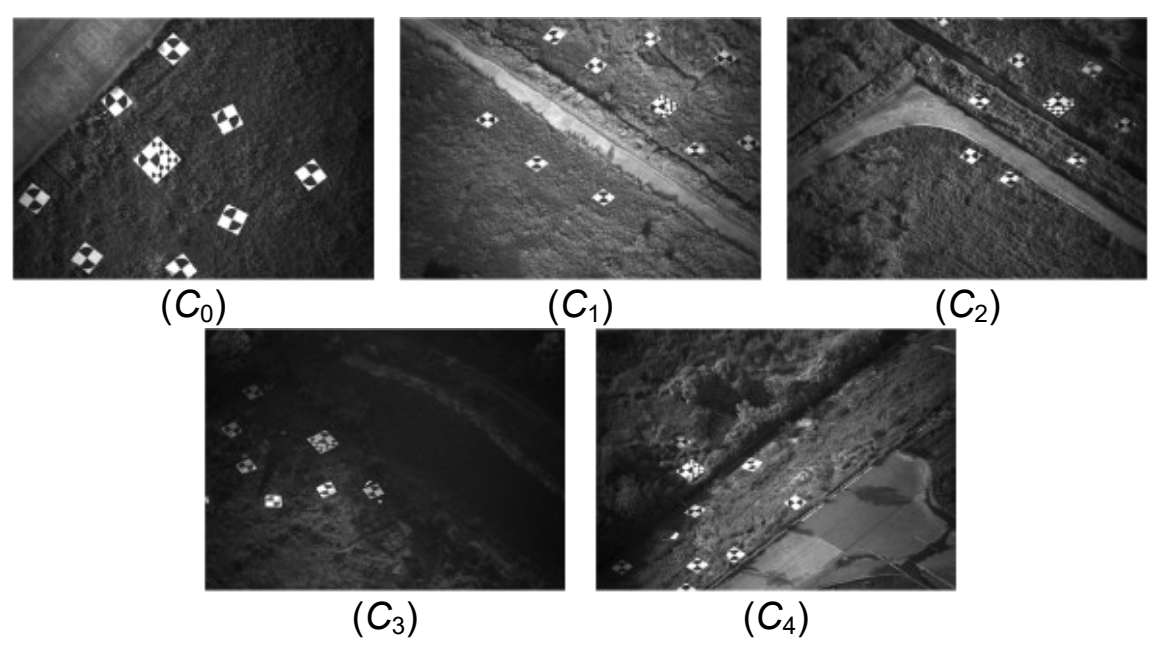

Fig. 11. Sample images obtained from the flight experiment.

## V. CONCLUSION

This paper presents a virtual-point-based framework to address the generalized absolute pose problem for multi-camera systems. By introducing the virtual point, we bridge the gap between standard and generalized camera models, enabling a unified pipeline that transforms existing PnP solvers into their generalized counterparts. Based on this formulation, we derive three competitive solvers (VGPc, VGPq, and VGPr) leveraging different rotation parameterizations. Experimental results demonstrate that the proposed solvers inherit the superior accuracy and efficiency of their original single-camera counterparts: VGPc achieves higher estimation accuracy, VGPq maintains certified global optimality, whereas VGPr delivers superior computational efficiency without accuracy degradation. The proposed framework provides efficient and robust solutions for pose estimation in multi-camera applications.

## Acknowledgment

The authors would like to thank the authors of DLSU [14], OPnP [8], SQPnP [13], UPnP [19], gDLS [20], gDLS+++ [22], and GAPS [23] for making their source code publicly available.